%% file: main.tex
\documentclass[a4paper]{spie}  %>>> use this instead for A4 paper
\usepackage{amsmath,amsfonts,amssymb}
\usepackage{graphicx}
\usepackage[colorlinks=true, allcolors=blue]{hyperref}

\usepackage{multirow}
\usepackage{booktabs,longtable,array}
\usepackage{subcaption}    % 子图排版（推荐）
\usepackage{xcolor} % 用于颜色定义
\definecolor{emptyColor}{RGB}{255, 255, 255}
\definecolor{floorColor}{RGB}{255, 120, 50}   % 地板颜色，对应 floor
\definecolor{wallColor}{RGB}{255, 192, 203}   % 粉红色，对应 wall
\definecolor{chairColor}{RGB}{255, 255, 0}    % 黄色，对应 chair
\definecolor{cabinetColor}{RGB}{0, 150, 245}  % 蓝色，对应 cabinet
\definecolor{doorColor}{RGB}{0, 255, 255}     % 青色，对应 door
\definecolor{tableColor}{RGB}{0, 175, 0}      % 绿色，对应 table
\definecolor{couchColor}{RGB}{255, 0, 0}      % 红色，对应 couch
\definecolor{shelfColor}{RGB}{127, 127, 127}  % 灰色，对应 shelf
\definecolor{windowColor}{RGB}{135, 60, 0}    % 棕色，对应 window
\definecolor{bedColor}{RGB}{160, 32, 240}     % 紫色，对应 bed
\definecolor{curtainColor}{RGB}{255, 0, 255}  % 粉红色，对应 curtain
\definecolor{refrigeratorColor}{RGB}{175, 0, 75}     % 棕色，对应 refrigerator
\definecolor{plantColor}{RGB}{200, 137, 137}  % 灰色，对应 plant
\definecolor{stairsColor}{RGB}{75, 0, 75}     % 紫色，对应 stairs
\definecolor{toiletColor}{RGB}{150, 240, 80}  % 绿色，对应 toilet

\definecolor{roofColor}{RGB}{255, 173, 50}   % 屋顶颜色，对应 roof
\definecolor{beamColor}{RGB}{100, 200, 0}    % 梁颜色，对应 beam
\definecolor{frameColor}{RGB}{100, 191, 255} % 框架颜色，对应 frame

\definecolor{exerciseEquipmentColor}{RGB}{0, 255, 127} % 健身器材颜色，对应 exercise equipment
\definecolor{microwaveColor}{RGB}{255, 200, 50} % 微波炉颜色，对应 microwave
\definecolor{printerColor}{RGB}{230, 230, 250} % 打印机颜色，对应 printer
\definecolor{ovenColor}{RGB}{214, 255, 50}   % 烤箱颜色，对应 oven

\definecolor{washbasinColor}{RGB}{50, 200, 192} % 洗手盆颜色，对应 washbasin
\definecolor{partitionColor}{RGB}{50, 55, 255} % 隔断颜色，对应 partition
\definecolor{pianoColor}{RGB}{150, 182, 255}  % 钢琴颜色，对应 piano
\definecolor{countertopColor}{RGB}{91, 50, 255} % 橱柜台面颜色，对应 countertop
\definecolor{drawerColor}{RGB}{200, 150, 45} % 抽屉颜色，对应 drawer
\definecolor{carpetColor}{RGB}{255, 100, 173}  % 地毯颜色，对应 carpet
\definecolor{bicycleColor}{RGB}{222, 150, 255} % 抽屉颜色，对应 bicycle
\definecolor{mailboxColor}{RGB}{255, 50, 173}  % 地毯颜色，对应 mailbox
\usepackage{comment}

\makeatletter
\newcommand{\eg}{\emph{e.g.}\@ifnextchar.{\!}{\ }}
\makeatother
\title{Exploring 2D backbone effects for indoor semantic occupancy prediction}

\author[a]{Shizhang Fang}
\author[a]{Wanling Ye}
\author[a]{Qi Zheng*}
\affil[a]{College of Electronics and Information Engineering, Shenzhen University, Shenzhen, China}
\authorinfo{Further author information: (Send correspondence to Qi Zheng.)\\
Qi Zheng: E-mail: qiz@szu.edu.cn, Telephone: 1 330 248 0490\\  
Shizhang Fang: E-mail: 2022280386@email.szu.edu.cn,  Telephone: 1 319 294 2539\\
Wanling Ye: E-mail: 2024280525@email.szu.edu.cn, Telephone: 1 895 948 7230\\
}

\begin{document} 
\maketitle

\begin{abstract}
Semantic occupancy prediction gives an embodied agent a voxel-level account of where space is free, occupied, and semantically meaningful. In RGB-D pipelines such as EmbodiedScan, the image encoder is often left as a default module, even though its features are the visual evidence later sampled into the 3D grid. We study this design choice directly. 

% \textcolor{red}{xxx.} Keeping the data processing, depth branch, projection rule, fusion neck, occupancy head, losses, optimizer, and training schedule unchanged, we replace only the 2D backbone.

A central finding is that changing the 2D backbone improves occupancy accuracy more than several carefully designed occupancy architectures or modules. We keep the main RGB-D projection, depth branch, and occupancy head fixed, and replace only the image backbone.
The compared encoders are CLIP-ResNet, CLIP-ViT, BLIP2, and DINOv2. Under the controlled setting, the measured mIoU changes substantially: DINOv2 obtains 30.55\%, BLIP2 obtains 29.49\%, CLIP-ViT obtains 24.33\%, and CLIP-ResNet obtains 17.41\%. The stronger encoders also exceed the original EmbodiedScan ResNet-50 baseline without modifying the downstream 3D fusion pipeline. Class-level results give a more detailed picture: DINOv2 is stronger on many layout and structural categories, whereas BLIP2 remains close on several object-centered classes. CLIP-ViT improves clearly over CLIP-ResNet, showing that the way CLIP features are exposed as dense tokens matters for voxel lifting. These results indicate that the image backbone is not a secondary engineering detail in embodied semantic occupancy, but a major source of variation in the final 3D prediction.

\end{abstract}


% \input{1_intro}
% \input{2_rel}
% \input{3_met}
% \input{4_exp}
% \input{5_conc}

\input{section/introduction}

\input{section/relatedwork}

\input{section/method}
\input{section/experiment}
\newpage
\input{section/conclusion}

% \acknowledgments
% We used OpenAI ChatGPT for language polishing, LaTeX formatting checks, and manuscript self-checking against the SPIE submission checklist. The prompts asked the tool to polish manuscript text, check formatting requirements, and identify possible LaTeX or submission-compliance issues. We reviewed and verified the technical content, experimental data, and conclusions.
\acknowledgments
We used OpenAI ChatGPT for language polishing, grammar refinement, LaTeX formatting verification, and manuscript self-checking against the SPIE submission guidelines. The tool was not used to generate experimental data, scientific claims, or research conclusions. All technical content, methodology, results, and interpretations were reviewed and verified by us.

Example prompts used with the tool included:
\begin{itemize}
    \item ``Polish the following paragraph for academic writing style while preserving the original technical meaning and terminology.''
    \item ``Check whether this manuscript section complies with SPIE formatting and submission requirements.''
    \item ``Review the following LaTeX code and identify possible compilation, formatting, or reference issues.''
    \item ``Identify grammatical errors, inconsistent notation, or unclear expressions in the following manuscript text.''
\end{itemize}

We take full responsibility for the accuracy, originality, and integrity of the submitted work.

% References
\bibliography{section/refs} % bibliography data in section/refs.bib
\bibliographystyle{spiebib} % makes bibtex use spiebib.bst

\end{document}

%% file: section/introduction.tex
% !TEX root = ./main.tex

\section{Introduction}
\label{sec:introduction}

% For an embodied agent, recognizing a chair in an image is not enough. The agent also needs to know where free space ends, which parts of the room are hidden, and how object labels are arranged in a shared 3D frame. Semantic occupancy prediction addresses this need by assigning occupancy and semantic labels to voxels. The resulting grid is a practical intermediate representation for navigation, manipulation, and scene memory because it keeps geometry and semantics in the same coordinate system.

For an embodied agent, recognizing a chair in an image is not enough. The agent also needs to know where free space ends, which parts of the room are hidden, and how object labels are arranged in a shared 3D frame. Indoor RGB-D scene understanding benchmarks and recent indoor occupancy studies make this requirement increasingly concrete~\cite{song2015sun}. Semantic occupancy prediction addresses this need by assigning occupancy and semantic labels to voxels. The resulting grid is a practical intermediate representation for navigation, manipulation, and scene memory because it keeps geometry and semantics in the same coordinate system.

The field has moved quickly, but along several different axes. TPVFormer~\cite{huang2023tri} and VoxFormer~\cite{li2023voxformer}%SparseOcc~\cite{tang2024sparseocc}, and COTR~\cite{ma2024cotr}
mainly ask how the 3D representation can be made cheaper or more effective. SelfOcc~\cite{huang2024selfocc} and OccCLIP~\cite{zhang2025occclip} look instead at supervision cost and semantic generalization. Indoor RGB-D scenes add a separate set of issues: furniture occludes other furniture, categories are unevenly distributed, and many objects are only partially seen from an ego-centric trajectory. EmbodiedScan~\cite{wang2024embodiedscan}, built from resources such as ScanNet, Matterport3D, and 3RScan, makes this indoor setting measurable.

In current semantic occupancy studies, most attention is placed on view transformation, voxel optimization, or multi-modal fusion architectures. Under this convention, the 2D image backbone is often treated as a pluggable and secondary engineering detail, and its choice usually follows a default setting rather than a controlled analysis.

This backbone-agnostic view is questionable in a visually dependent pipeline such as EmbodiedScan. Each voxel receives semantic evidence by projecting its center onto the image feature map and sampling the corresponding 2D descriptor. Therefore, the alignment quality and class separability of the image features constrain the prediction before 3D fusion takes place. If the initial 2D features fail to preserve fine-grained geometric boundaries, increasing the complexity of the later 3D decoder may not recover the lost spatial-semantic information.

This motivates a direct test of the backbone-agnostic view. If the 2D representation already limits the quality of voxel-level semantic evidence, then backbone choice should be treated as a core experimental variable rather than a default implementation detail.

% Against this background, one design choice is easy to overlook. Occupancy papers usually discuss view transformation, voxelization, fusion, or decoder capacity; the 2D image backbone often appears as a swappable component.
% In the EmbodiedScan RGB-D pipeline, however, each voxel obtains part of its semantic evidence by projecting the voxel center into the image feature map and sampling the corresponding 2D descriptor. Those descriptors are averaged across valid views and fused with depth-derived geometry. If the image features are poorly aligned or weakly separable, the error is already present before the 3D head starts to reason over the volume.

% In the EmbodiedScan RGB-D pipeline, however, the image feature is the semantic signal that is projected into the voxel grid.It is sampled at projected grid locations, averaged across views, and then fused with depth-derived geometry. A poor image representation can therefore enter the model before the 3D head has any chance to correct it.

This issue becomes more visible now that pretrained visual encoders are no longer variants of the same recipe. CLIP~\cite{radford2021learning} brings image-language alignment; BLIP2~\cite{li2023blip2} uses a frozen visual encoder inside a vision-language system; DINOv2~\cite{oquab2023dinov2} is trained self-supervised and is often used for dense visual correspondence. These objectives do not encourage the same spatial behavior. A representation that separates image-level concepts may not preserve object boundaries after projection, while a locally stable representation may be easier to reuse in voxel prediction.

Our study isolates this single variable. We do not introduce a new decoder or a new fusion block. Instead, we keep the RGB-D data pipeline, point branch, voxel projection, fusion neck, occupancy head, losses, optimizer, schedule, and evaluation protocol fixed, and change only the 2D encoder. With this setup, differences in mIoU are mainly differences in the visual representation that is lifted from image space into the 3D grid.

Concretely, we compare four pretrained backbones: CLIP-ResNet, CLIP-ViT, BLIP2, and DINOv2. The first two compare different visual architectures under CLIP-style language supervision, BLIP2 represents vision-language pretraining with a frozen visual encoder, and DINOv2 represents large-scale self-supervised visual pretraining. All variants are adapted to the same FPN-compatible interface and inserted into the same projection-based RGB-D occupancy model.

% The results are not a small implementation detail. DINOv2 reaches 30.55\% mIoU, BLIP2 reaches 29.49\%, CLIP-ViT reaches 24.33\%, and CLIP-ResNet reaches 17.41\%. The CLIP-ViT result is much better than CLIP-ResNet, so the visual architecture matters even under a related pretraining family. The class-wise results also split the story: DINOv2 is more reliable on many structural and layout-related categories, while BLIP2 remains close on several object-centered categories.
The measured effect is larger than expected for a front-end replacement. DINOv2 reaches 30.55\% mIoU, BLIP2 reaches 29.49\%, CLIP-ViT reaches 24.33\%, and CLIP-ResNet reaches 17.41\%. This gap is larger than the improvement from several occupancy-specific module changes reported on the same benchmark. The CLIP-ViT result is also much higher than CLIP-ResNet, so visual architecture matters even inside a related pretraining family. Class-wise results split the story further: DINOv2 is more reliable on many structural and layout-related categories, while BLIP2 remains close on several object-centered categories.

The main contributions of this paper are:
\begin{itemize}
    \item We present a controlled study of image backbone effects for EmbodiedScan semantic occupancy prediction, isolating the 2D encoder while fixing the RGB-D occupancy pipeline.
    \item We compare CLIP-ResNet, CLIP-ViT, BLIP2, and DINOv2 under the same projection-based 3D fusion framework, and show that DINOv2 gives the best overall transfer while BLIP2 remains close.
    \item We benchmark the backbone variants against the original EmbodiedScan result, our EmbodiedScan reproduction, DROcc(SwinU), and DROcc, showing that the 2D backbone alone can exceed several task-specific architectural baselines.
    \item We analyze both common categories and the 15 least frequent categories, finding that stronger backbones improve rare-object prediction but do not remove the long-tail weakness of indoor occupancy.
\end{itemize}
% \begin{itemize}
%     \item We present a controlled study of image backbone effects for EmbodiedScan semantic occupancy prediction, isolating the 2D encoder while fixing the RGB-D occupancy pipeline.
%     \item We compare CLIP-ResNet, CLIP-ViT, BLIP2, and DINOv2 under the same projection-based 3D fusion framework.
%     \item We show that DINOv2 achieves the best overall transfer, BLIP2 remains close and competitive on several object categories, CLIP-ViT improves over CLIP-ResNet, and CLIP-ResNet is substantially weaker in this dense voxel prediction setting.
%     \item We provide class-wise observations showing that image backbone choice affects both average mIoU and category-level behavior.
% \end{itemize}

%% file: section/relatedwork.tex
% !TEX root = ./main.tex

\section{Related Work}
\label{sec:related_work}

\noindent\textbf{Semantic occupancy prediction.}
Semantic occupancy prediction inherits the goal of semantic scene completion: a partial observation must be converted into a voxel-level description of geometry and semantics. Recent work has improved this task by changing how views are transformed, how voxels are represented, how supervision is reduced~\cite{huang2024selfocc}, or how semantic priors are introduced~\cite{zhang2025occclip}. After August 2025, occupancy research has also moved toward causal 2D-to-3D lifting~\cite{chen2025semanticcausalocc}, vertical-slice representations~\cite{huang2025slicesemocc}, pseudo-label supervision~\cite{hayes2025easyocc}, temporal world modeling~\cite{jin2025occtens}, implicit occupancy supervision for vision-language-action models~\cite{liu2025occvla}, LiDAR-free native 3D supervision~\cite{boeder2025shelfocc}, and generalized urban occupancy~\cite{cao2026occany}. These directions are important, but they mainly alter the 3D representation or the decoder side of the pipeline. Our work asks a complementary question: before those 3D modules operate, how much does the sampled 2D visual representation already determine the final occupancy result?
% Semantic occupancy prediction inherits the goal of semantic scene completion: a partial observation must be converted into a voxel-level description of geometry and semantics. Recent work has improved this task by changing how views are transformed, how voxels are represented, how supervision is reduced~\cite{huang2024selfocc}, or how semantic priors are introduced~\cite{zhang2025occclip}. These directions are important, but they mainly alter the 3D representation or the decoder side of the pipeline. Our work asks a complementary question: before those 3D modules operate, how much does the sampled 2D visual representation already determine the final occupancy result?

\noindent\textbf{Indoor embodied occupancy.}
Indoor embodied scenes are different from outdoor driving scenes. They are cluttered, object categories are long-tailed, and RGB-D observations are collected from ego-centric viewpoints. EmbodiedScan~\cite{wang2024embodiedscan} provides the benchmark setting used in this paper, and its RGB-D baseline follows a projection-based design: image features are sampled into a voxel grid, depth features provide geometric evidence, and both volumes are fused for occupancy prediction. DROcc~\cite{fang2026drocc} improves indoor RGB-D occupancy with a stronger fusion architecture. In contrast, we keep the EmbodiedScan-style pipeline fixed and study whether the image backbone itself changes the dense 3D prediction.

\noindent\textbf{Image backbones for projected 3D features.}
Projection-based 3D perception does not use image features only for image-level classification. The feature map is sampled at projected voxel locations, moved into 3D, and fused with geometry. This makes the visual encoder a more consequential choice than it may appear in a modular implementation. Our comparison covers a convolutional CLIP variant based on ResNet~\cite{he2016deep,radford2021learning}, a CLIP-ViT variant, the frozen visual encoder used in BLIP2~\cite{li2023blip2}, and DINOv2~\cite{oquab2023dinov2}. The goal is not to introduce another occupancy decoder, but to measure how these pretrained visual representations behave under the same RGB-D lifting and fusion pipeline.

%% file: section/method.tex
% !TEX root = ./main.tex

\section{Method}
\label{sec:method}

This section fixes the experimental ground on which the backbone comparison is made. We use the same RGB-D occupancy pipeline for all runs and change only the image encoder. The point branch, projection operation, fusion neck, occupancy head, losses, optimizer, and evaluator are kept in place, so the comparison is about the visual features that enter the voxel grid rather than about a new 3D architecture.

\subsection{Pipeline Overview}
\label{sec:pipeline_overview}

\begin{figure*}[t]
\centering
\begin{subfigure}[t]{1\linewidth}
    \centering
    \includegraphics[width=\linewidth]{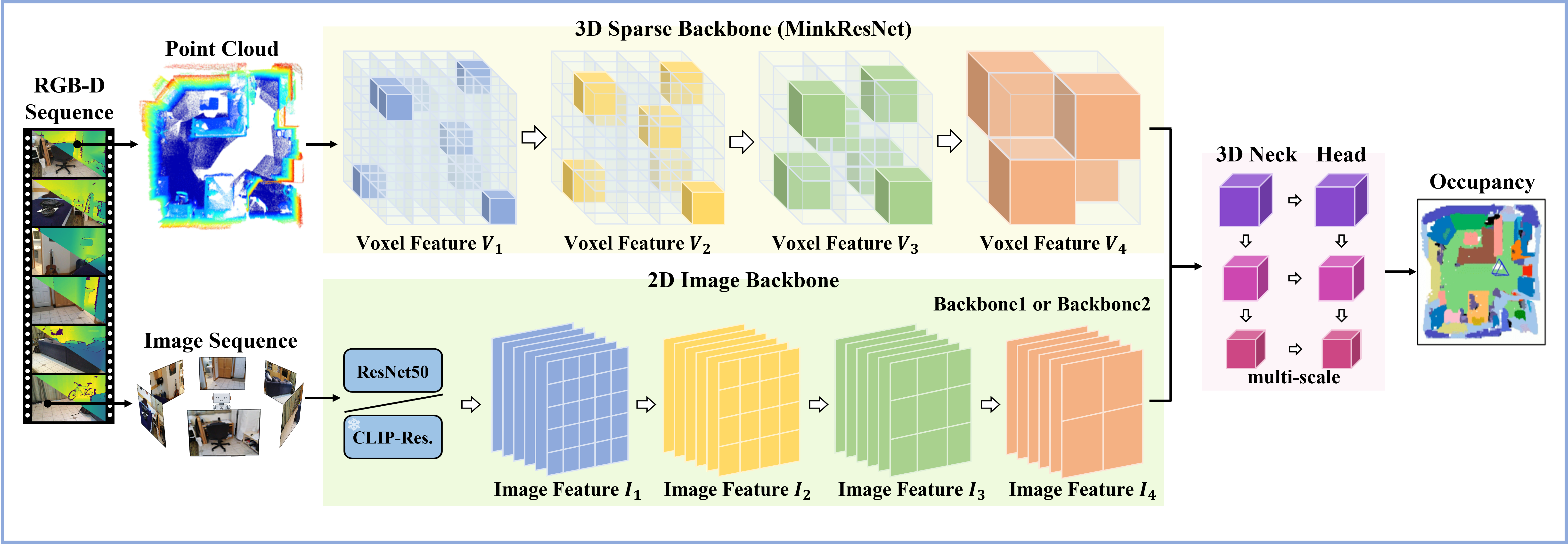}
    \caption{Fixed RGB-D occupancy pipeline}
    \label{fig:pipeline_main}
\end{subfigure}

\vspace{0.5em}

\begin{subfigure}[t]{1\linewidth}
    \centering
    \includegraphics[width=\linewidth]{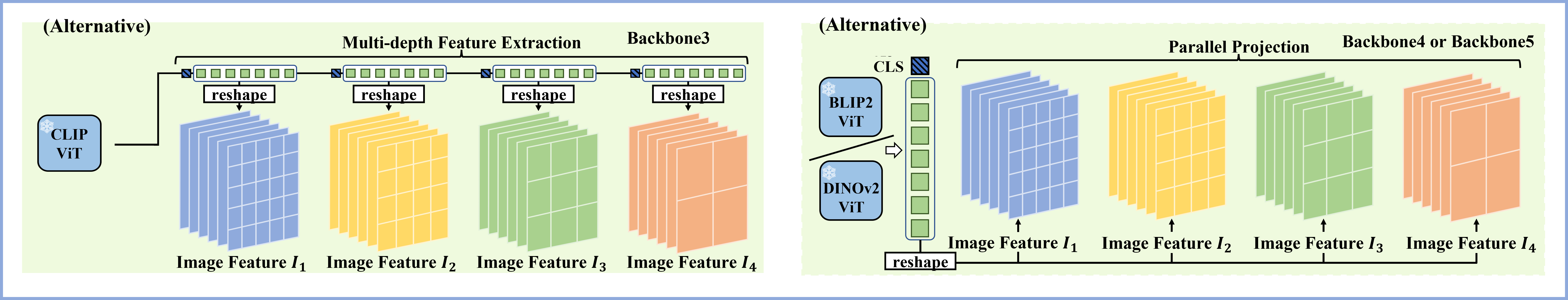}
    \caption{Alternative image backbones}
    \label{fig:pipeline_backbone}
\end{subfigure}
% \caption{Controlled pipeline used for the backbone comparison. (a) The RGB branch produces an FPN-compatible 2D feature hierarchy, and the depth branch produces a sparse-to-dense 3D feature volume. For both branches, feature resolutions decrease from left to right. Voxel centers are projected to the selected image feature map, the corresponding descriptors are sampled, and the sampled descriptors are written back to their original 3D grid cells before RGB-D fusion. (b) The experimental intervention replaces only the image backbone with CLIP-ResNet, CLIP-ViT, BLIP2, or DINOv2. The projection rule, depth branch, fusion neck, occupancy head, losses, and training protocol are kept fixed.}
\caption{Controlled pipeline used for the backbone comparison. (a) The RGB branch produces an FPN-compatible 2D feature hierarchy, and the depth branch produces a sparse-to-dense 3D feature volume. For both branches, feature resolutions decrease from left to right. Voxel centers or sparse voxel coordinates are projected to the selected image feature map, the corresponding descriptors are sampled, and the sampled descriptors are written back to the same 3D grid coordinates before RGB-D fusion. The projection step uses visibility and image-boundary masks but does not introduce an additional occlusion reasoning module. (b) The experimental intervention replaces only the image backbone with CLIP-ResNet, CLIP-ViT, BLIP2, or DINOv2. For transformer backbones, class tokens are discarded and patch tokens are reshaped into spatial maps before FPN adaptation. The projection rule, depth branch, fusion neck, occupancy head, losses, and training protocol are kept fixed.}

\label{fig:model_pipeline}
\end{figure*}

Figure~\ref{fig:model_pipeline} shows the pipeline used in every experiment. A sample contains several posed RGB-D views. The RGB branch sends each image through a replaceable 2D backbone and an FPN. For voxel lifting, the model projects 3D grid centers onto the 2D feature map, samples the descriptors at those projected locations, and writes the sampled descriptors back to the corresponding grid cells. The depth branch follows a separate route: depth maps are converted to scene points, voxelized, and encoded by a Minkowski ResNet-34. The resulting image and depth volumes are concatenated before the 3D neck and occupancy head.

% \begin{figure*}[t]
%     \centering
%     % 第一张子图
%     \begin{subfigure}[c]{1\textwidth}
%         \centering
%         \includegraphics[width=\linewidth]{./pipeline1.png}
%         \caption{RGB-D semantic occupancy pipeline}
%     \label{fig:model_pipeline}
%     \end{subfigure}

%     \vspace{3pt}
    
%     % 第二章子图
%     \begin{subfigure}[c]{1\textwidth}
%         \centering
%         \includegraphics[width=\textwidth]{./alternative_backbone.png}
%         \caption{Alternative 2D Image backbone}
%         \label{fig:alternative_backbone}
%     \end{subfigure}

%     \vspace{3pt}
    
%     \caption{RGB-D semantic occupancy pipeline used for the backbone comparison. The RGB branch produces an FPN-compatible 2D feature hierarchy; transformer backbones discard the class token and reshape patch tokens into spatial maps before FPN adaptation. Both the 2D features and 3D point features form multi-scale pyramids with lower spatial resolution from left to right. Voxel centers are projected to the selected image feature map, sampled at the projected locations, and assigned back to their original 3D grid cells. The lifted image volume is then fused with the depth-derived feature volume for occupancy prediction.}
% \end{figure*}

\subsection{Problem Formulation}
\label{sec:problem_formulation}

Given a scene observed by $V$ posed RGB-D views, each sample contains RGB images $\{I_v\}_{v=1}^{V}$, depth observations $\{D_v\}_{v=1}^{V}$, camera intrinsics $\{K_v\}_{v=1}^{V}$, and camera-to-scene transformations. The target is a semantic occupancy grid
\begin{equation}
    Y \in \{0, 1, \ldots, C\}^{X \times Y \times Z},
\end{equation}
where each voxel is assigned either empty space or one of $C$ semantic object categories. We follow the EmbodiedScan occupancy setting~\cite{wang2024embodiedscan}; the concrete grid size, scene range, and category count are reported with the experimental protocol in Section~\ref{sec:dataset_eval}.

The central question of this work is how the 2D visual representation affects the final 3D occupancy prediction. Let $\phi$ denote the image backbone and let the remaining occupancy network be denoted by $f_{\theta}$. The prediction can be written abstractly as
\begin{equation}
    \hat{Y} = f_{\theta}(\phi(I_{1:V}), D_{1:V}, K_{1:V}, T_{1:V}),
\end{equation}
where $D_{1:V}$ are depth observations and $T_{1:V}$ are camera poses. We compare different choices of $\phi$ while keeping $f_{\theta}$, data processing, voxelization, training schedule, and losses unchanged. This isolates the role of the transferred image representation from other sources of architectural or optimization variation.

\subsection{Fixed RGB-D Occupancy Pipeline}
\label{sec:fixed_pipeline}

We build on the multi-view RGB-D occupancy pipeline of EmbodiedScan~\cite{wang2024embodiedscan}. Each training sample uses multiple posed RGB-D views. RGB images are normalized before feature extraction. Depth maps are converted into point clouds, transformed into a shared scene coordinate frame, range filtered, and sampled to a fixed number of points. The occupancy target is represented in the same metric scene frame as the voxel grid.

\paragraph{Image feature extraction.}
% Each RGB view is passed through the same 2D image encoder and feature pyramid network (FPN)~\cite{lin2017feature}. The backbone output is converted to the channel layout expected by the FPN. We use the highest-resolution FPN level for lifting, because the sampling operation is tied to projected voxel centers and loses detail if the image map is too coarse.
Each RGB view is passed through the same 2D image encoder and feature pyramid network (FPN)~\cite{lin2017feature}. The backbone output is converted to the channel layout expected by the FPN. For ResNet-style backbones, the native convolutional stages provide the feature hierarchy. For ViT-style backbones, we discard the class token, reshape the remaining patch tokens according to the image patch grid, apply linear or $1\times1$ projections to match the FPN channel dimensions, and resize the resulting spatial maps to the same feature scales used by the baseline FPN. We use the highest-resolution FPN level for lifting, because the sampling operation is tied to projected voxel centers and loses detail if the image map is too coarse. In the sparse stage, the same calibrated projection is applied to the coordinates of the sparse voxel set $V_4$; sampled image descriptors are indexed back by their original voxel coordinates, which keeps $I_4$ and $V_4$ aligned in the shared scene grid.

\paragraph{Projection-based voxel lifting.}
For each voxel center $x_g$ in the target 3D grid, the model first projects the grid point onto the 2D feature map of every selected camera view using the known camera parameters:
\begin{equation}
    u_{v,g} = \pi(K_v T_v x_g),
\end{equation}
where $\pi(\cdot)$ is the perspective projection. A valid projected point is used to bilinearly sample the 2D feature map. The sampled descriptor is not a new 3D point; it is written back to the voxel center that produced the projection. Thus, in implementation terms, the operation is closer to ``project a grid point, sample an image descriptor, and store it at the same grid point''. For multiple views, we average only the valid samples:
\begin{equation}
    F^{img}_g =
    \frac{\sum_{v=1}^{V} m_{v,g} \, S(F_v, u_{v,g})}
    {\max(\sum_{v=1}^{V} m_{v,g}, 1)},
\end{equation}
% Here $F_v$ is the feature map for view $v$, $S(\cdot)$ denotes bilinear sampling, and $m_{v,g}$ marks whether grid cell $g$ has a valid projection in that view. Because this step copies image descriptors into the volume before 3D fusion, errors in local alignment or semantic separability can directly affect the final occupancy logits.

Here $F_v$ is the feature map for view $v$, $S(\cdot)$ denotes bilinear sampling, and $m_{v,g}$ marks whether grid cell $g$ has a valid projection in that view. The mask checks whether the projected point lies inside the image and has a valid camera projection. It does not perform explicit z-buffering or depth-order reasoning among voxels that fall on the same image pixel. Therefore, an occluded voxel may still receive an appearance descriptor from a visible surface along the same ray. We keep this lifting rule unchanged for all backbones, so the comparison remains controlled, but this limitation explains why the 2D encoder is not the only bottleneck for categories with weak observability. Because this step copies image descriptors into the volume before 3D fusion, errors in local alignment, visibility, or semantic separability can directly affect the final occupancy logits.

\paragraph{Depth branch and dense fusion.}
The point branch uses the aggregated depth points as a geometric input. Points are voxelized into a sparse tensor and processed by a Minkowski ResNet-34~\cite{choy20194d}. The final sparse feature level is densified into the target voxel grid, producing a depth feature volume $F^{dep}$. The image and depth volumes are concatenated channel-wise:
\begin{equation}
    F^{fuse} = [F^{img}; F^{dep}],
\end{equation}
and processed by a 3D neck followed by an occupancy head. The resulting logits are
\begin{equation}
    Z = h_{\theta}(F^{fuse}).
\end{equation}
This fusion design gives the model complementary cues: depth supplies metric geometry and visible surface structure, while image features supply appearance and semantic evidence for class assignment.

\subsection{Training Objective}
\label{sec:training_objective}

The occupancy head is supervised with voxel-wise semantic labels. Invisible voxels are assigned the ignore label 255 and are excluded from cross-entropy. Following the baseline implementation, the total loss combines semantic cross-entropy, semantic scaling loss, and geometric scaling loss:
\begin{equation}
    \mathcal{L} =
    \sum_{s} \lambda_s
    \left(
    \mathcal{L}^{s}_{ce}
    + \mathcal{L}^{s}_{sem}
    + \mathcal{L}^{s}_{geo}
    \right),
\end{equation}
where $s$ indexes the supervision scale and $\lambda_s = 0.5^s$ down-weights lower-resolution predictions. Cross-entropy optimizes the voxel-wise semantic decision, semantic scaling encourages class-level consistency, and geometric scaling strengthens the occupied-versus-empty structure. These losses match the fixed occupancy baseline and are not changed across backbone variants.

\subsection{Controlled Backbone Intervention}
\label{sec:backbone_intervention}

For the backbone study, only $\phi$ is replaced. This restriction matters because occupancy scores are otherwise sensitive to voxel size, point sampling, depth quality, decoder capacity, and loss weights. Holding these choices fixed keeps the experiment focused on the image representation that is copied into the grid.

We evaluate four pretrained visual backbones.
\begin{itemize}
    \item \textbf{CLIP-ResNet.} This variant uses a ResNet-style image encoder initialized from CLIP-pretrained weights~\cite{he2016deep,radford2021learning}. CLIP pretraining aligns image representations with language supervision and often produces strong image-level semantic discrimination. In our setting, it tests whether such global semantic alignment transfers effectively to dense voxel prediction.
    \item \textbf{CLIP-ViT.} This variant uses a CLIP-pretrained Vision Transformer visual trunk~\cite{radford2021learning}. Patch tokens from intermediate transformer layers are reshaped into spatial feature maps and projected to the FPN channel dimensions. This model tests whether CLIP-style language supervision becomes more suitable for voxel lifting when the visual encoder exposes dense token features rather than a convolutional hierarchy.
    \item \textbf{BLIP2 visual backbone.} This variant uses the frozen visual encoder from BLIP2~\cite{li2023blip2}. Its features are adapted into FPN-compatible spatial maps and channel dimensions. BLIP2 provides vision-language representations shaped by image-text and query-based modeling, making it a useful comparison for object-centric semantic transfer.
    \item \textbf{DINOv2 backbone.} This variant uses a DINOv2 Vision Transformer~\cite{oquab2023dinov2}. Patch tokens are reshaped into 2D feature maps, projected to the channel dimensions expected by FPN, and resized to the required feature scales. DINOv2 is trained with self-supervised objectives that are known to preserve strong visual structure and dense correspondence cues, which are especially relevant to projection-based 3D lifting.
\end{itemize}

% All four variants output feature levels compatible with the same downstream occupancy model. The adapters only make the feature tensor shapes consistent; they do not change the voxel lifting rule, 3D fusion module, or supervision. Therefore, the comparison measures how different pretraining paradigms and visual architectures behave after being inserted into the same geometric occupancy pipeline.
All four variants output feature levels compatible with the same downstream occupancy model. The adapters only make the feature tensor shapes consistent; they do not change the voxel lifting rule, 3D fusion module, or supervision. In Fig.~\ref{fig:model_pipeline}, Backbone 3 denotes the CLIP-ViT setting, where dense patch tokens come from CLIP language-aligned pretraining. Backbones 4 and 5 denote BLIP2 and DINOv2, respectively; they use the same token-to-map and FPN adaptation interface, but differ in the pretrained visual encoder and learning objective. This design keeps the adapter interface fixed while testing whether language-aligned, vision-language, and self-supervised visual representations behave differently after projection into 3D. Therefore, the comparison measures how different pretraining paradigms and visual architectures behave after being inserted into the same geometric occupancy pipeline.

\subsection{Evaluation Principle}
\label{sec:evaluation_principle}

All variants are evaluated with the same semantic occupancy protocol. We report mIoU and class-wise IoU because the average alone hides much of the behavior in indoor scenes. Large structural classes such as floor and wall are observed often, while small furniture and appliances appear sparsely. A useful backbone should therefore help the mean score without collapsing the less frequent categories.

%% file: section/experiment.tex
% !TEX root = ./main.tex

\section{Experiments}
\label{sec:experiments}

This section evaluates how different image backbones affect EmbodiedScan semantic occupancy prediction. All backbone variants are tested under the RGB-D occupancy pipeline described in Section~\ref{sec:method}. The purpose of the experiments is therefore not to compare different decoders, losses, or input modalities, but to isolate the contribution of the 2D image representation that is lifted into the 3D voxel volume.

\subsection{Dataset and Evaluation Protocol}
\label{sec:dataset_eval}

Experiments are conducted on the EmbodiedScan semantic occupancy benchmark~\cite{wang2024embodiedscan}. Each scene is represented by posed RGB-D observations and evaluated on a dense voxel grid. Following the original occupancy setting, the model predicts a $40 \times 40 \times 16$ grid with 81 channels, corresponding to empty space and 80 semantic object categories. We report mean Intersection-over-Union (mIoU) over all semantic categories and further analyze class-wise IoU values.

The main comparison is controlled across four image backbones: CLIP-ResNet, CLIP-ViT, BLIP2, and DINOv2. The same data split, RGB-D inputs, 3D branch, fusion head, training objective, and evaluation metric are used for all variants. This protocol makes the measured differences attributable mainly to the image backbone and its transferred feature representation.

\subsection{Implementation Details}
\label{sec:implementation}

All variants use the same multi-view RGB-D setting. During training, ten RGB-D views are sampled from each scene. RGB images are resized to $480 \times 480$ and normalized before being processed by the image backbone and FPN. The point branch uses Minkowski ResNet-34~\cite{choy20194d}. The fusion neck receives a 256-channel image volume and a 512-channel depth volume after projection and sparse-to-dense conversion.

All models are trained for 24 epochs using AdamW with a learning rate of $1\times10^{-4}$ and weight decay of $1\times10^{-2}$. Training uses four GPUs with one sample per GPU. The same optimizer, schedule, data pipeline, loss functions, and evaluator are used for all backbone variants.

% \begin{table*}[t]
% \centering
% \caption{Backbone comparison on EmbodiedScan semantic occupancy prediction. Values are IoU in percentage. The best result in each column is shown in bold. ``refri.'' denotes refrigerator.}
% \label{tab:backbone_main}
% \resizebox{\linewidth}{!}{
% \begin{tabular}{l|c|cccccccccccccccc}
% \toprule
% Backbone & mIoU & empty & floor & wall & chair & cabinet & door & table & couch & shelf & window & bed & curtain & refri. & plant & stairs & toilet \\
% \midrule
% CLIP-ResNet & 17.41 & 73.84 & 70.04 & 51.34 & 51.93 & 24.66 & 24.24 & 40.88 & 44.74 & 38.95 & 25.74 & 48.76 & 41.90 & 13.94 & 25.29 & 35.10 & 50.47 \\
% CLIP-ViT & 24.33 & 74.67 & 70.90 & 54.76 & 55.14 & 31.07 & 33.69 & 43.24 & 51.76 & 42.52 & 29.97 & 55.53 & 51.22 & 31.16 & 40.24 & 38.66 & 64.54 \\
% BLIP2 & 29.49 & 74.22 & 70.33 & 55.64 & \textbf{57.99} & 34.51 & 40.33 & \textbf{50.12} & \textbf{58.70} & \textbf{43.75} & 31.81 & 58.87 & 54.29 & 38.20 & \textbf{41.10} & \textbf{43.13} & 66.43 \\
% DINOv2 & \textbf{30.55} & \textbf{75.03} & \textbf{71.26} & \textbf{57.10} & 57.98 & \textbf{34.95} & \textbf{41.09} & 47.75 & 58.59 & 43.72 & \textbf{34.40} & \textbf{62.83} & \textbf{55.70} & \textbf{39.80} & 40.46 & 37.00 & \textbf{67.52} \\
% \bottomrule
% \end{tabular}
% }
% \end{table*}

\begin{table*}[t]
\centering
\caption{RGB-D semantic occupancy comparison on EmbodiedScan. Values are IoU in percentage. ``Baseline'' denotes EmbodiedScan. ``Rep.'' is our reproduction of the EmbodiedScan RGB-D baseline. ``b.'' denotes replacing only the image backbone in the fixed RGB-D pipeline. ``refri.'' denotes refrigerator.}
\label{tab:backbone_main}
\resizebox{\linewidth}{!}{
\begin{tabular}{l|c|cccccccccccccccc}
\toprule
Method & mIoU & empty & floor & wall & chair & cabinet & door & table & couch & shelf & window & bed & curtain & refri. & plant & stairs & toilet \\
\midrule
Baseline~\cite{wang2024embodiedscan} & 19.97 & 71.21 & 64.92 & 55.00 & 52.04 & 27.35 & 33.97 & 47.93 & 46.26 & 31.87 & 27.98 & 46.58 & 46.56 & 24.05 & 39.01 & 24.40 & 67.79 \\
Baseline (rep.)~\cite{wang2024embodiedscan} & 21.20 & 73.55 & 69.61 & 53.33 & 52.64 & 30.11 & 31.37 & 42.49 & 46.46 & 41.92 & 28.00 & 49.30 & 47.00 & 19.60 & 36.75 & 32.52 & 62.46 \\
DROcc(SwinU)~\cite{fang2026drocc} & 22.14 & 73.74 & 69.52 & 52.70 & 50.89 & 27.39 & 29.22 & 38.42 & 45.04 & 40.00 & 26.26 & 49.47 & 45.99 & 18.96 & 36.35 & 33.22 & 60.18 \\
DROcc~\cite{fang2026drocc} & 23.38 & 73.46 & 70.96 & 52.99 & 50.06 & 30.15 & 31.24 & 40.70 & 47.33 & 41.12 & 26.88 & 51.79 & 49.40 & 24.49 & 35.08 & 36.37 & 61.48 \\
\midrule\midrule
b.CLIP-ResNet & 17.41 & 73.84 & 70.04 & 51.34 & 51.93 & 24.66 & 24.24 & 40.88 & 44.74 & 38.95 & 25.74 & 48.76 & 41.90 & 13.94 & 25.29 & 35.10 & 50.47 \\
b.CLIP-ViT & 24.33 & 74.67 & 70.90 & 54.76 & 55.14 & 31.07 & 33.69 & 43.24 & 51.76 & 42.52 & 29.97 & 55.53 & 51.22 & 31.16 & 40.24 & 38.66 & 64.54 \\
b.BLIP2 & 29.49 & 74.22 & 70.33 & 55.64 & \textbf{57.99} & 34.51 & 40.33 & \textbf{50.12} & \textbf{58.70} & \textbf{43.75} & 31.81 & 58.87 & 54.29 & 38.20 & \textbf{41.10} & \textbf{43.13} & 66.43 \\
b.DINOv2 & \textbf{30.55} & \textbf{75.03} & \textbf{71.26} & \textbf{57.10} & 57.98 & \textbf{34.95} & \textbf{41.09} & 47.75 & 58.59 & 43.72 & \textbf{34.40} & \textbf{62.83} & \textbf{55.70} & \textbf{39.80} & 40.46 & 37.00 & \textbf{67.52} \\
\bottomrule
\end{tabular}
}
\end{table*}

%\begin{table}[t]
%\centering
%\caption{Reference RGB-D occupancy performance on EmbodiedScan. EmbodiedScan~\cite{wang2024embodiedscan} and DROcc~\cite{fang2026drocc} provide scale for the benchmark, while our main controlled comparison is the backbone study in Table~\ref{tab:backbone_main}.}
%\label{tab:reference_results}
%\begin{tabular}{l|c}
%\toprule
%Method & mIoU (\%) \\
%\midrule
%EmbodiedScan ResNet-50 baseline & 19.97 \\
%DROcc (SwinU) & 22.14 \\
%DROcc & 23.38 \\
%Ours with CLIP-ResNet backbone & 17.41 \\
%Ours with CLIP-ViT backbone & 24.33 \\
%Ours with BLIP2 backbone & 29.49 \\
%Ours with DINOv2 backbone & \textbf{30.55} \\
%\bottomrule
%\end{tabular}
%\end{table}

\subsection{Main Quantitative Results}
\label{sec:main_results}

Table~\ref{tab:backbone_main} gives the main comparison. With all downstream 3D components fixed, DINOv2 reaches 30.55\% mIoU and BLIP2 reaches 29.49\%. CLIP-ViT is lower at 24.33\%, but it still improves over CLIP-ResNet by 6.92 points. The spread from CLIP-ResNet to DINOv2 is 13.14 points, which is too large to treat the image encoder as a minor setting. Because the point branch, fusion module, losses, and schedule are unchanged, the result points to the transferred 2D representation itself.

% The reference results in Table~\ref{tab:reference_results} provide additional context. The original EmbodiedScan RGB-D baseline with a ResNet-50 image backbone reports 19.97\% mIoU, while DROcc reports 23.38\% mIoU. Under our controlled backbone replacement setting, CLIP-ResNet reaches 17.41\%, CLIP-ViT reaches 24.33\%, BLIP2 reaches 29.49\%, and DINOv2 reaches 30.55\% mIoU. The comparison with the original ResNet-50 baseline is useful because it shows that replacing the image encoder can either hurt or help: the tested CLIP-ResNet is weaker than the baseline, while CLIP-ViT, BLIP2, and DINOv2 exceed it without changing the downstream RGB-D occupancy pipeline. 
Table~\ref{tab:backbone_main} shows that the 2D backbone is a large source of variation. Our reproduced EmbodiedScan baseline reaches 21.20\% mIoU, close to but slightly above the reported 19.97\%. DROcc reaches 23.38\%, and DROcc(SwinU) reaches 22.14\%. In the same benchmark setting, replacing only the image backbone gives 24.33\% with CLIP-ViT, 29.49\% with BLIP2, and 30.55\% with DINOv2. Thus, the improvement from a stronger image backbone is larger than the gain from several occupancy-specific module changes represented by these reference rows.

The comparison is not uniformly positive. CLIP-ResNet falls to 17.41\%, below both the reported and reproduced EmbodiedScan baselines. This negative case is useful: it shows that pretrained semantic alignment by itself is not enough. The features must also remain spatially useful after projection, view aggregation, and voxel-level decoding. CLIP-ViT improves every common category over CLIP-ResNet, indicating that dense patch tokens are easier to reuse than the tested ResNet-style CLIP hierarchy. BLIP2 and DINOv2 then give a much larger jump, with DINOv2 producing the best overall mIoU.
This does not change the focus of the paper into a decoder comparison; instead, it clarifies that backbone selection alone can produce large performance shifts.

\subsection{Class-Wise Analysis}
\label{sec:classwise}

DINOv2 gives the best IoU on 10 of the 16 commonly reported categories: empty, floor, wall, cabinet, door, window, bed, curtain, refrigerator, and toilet. Many of these categories involve room layout, broad surfaces, or stable spatial boundaries. This pattern matches the expectation that DINOv2 features preserve local visual structure after voxel lifting.

BLIP2 has a different profile. It is best on chair, table, couch, shelf, plant, and stairs. These categories are more object-centered, and their appearance can be more important than large-scale layout. The result therefore does not reduce to a single ranking of backbones: DINOv2 is steadier for structural categories, while BLIP2 remains competitive for object-level semantics. CLIP-ViT lies between CLIP-ResNet and the two strongest variants, confirming that tokenized visual features help but do not fully close the gap to BLIP2 or DINOv2.

\begin{table*}[t]
\centering
\caption{IoU on the 15 least frequent classes used for long-tail analysis. Values are percentages. ``exer.'' denotes exercise equipment. These categories follow the rare-class grouping used in DROcc~\cite{fang2026drocc}.}
\label{tab:longtail}
\resizebox{\linewidth}{!}{
\begin{tabular}{l|c|ccccccccccccccc}
\toprule
Method & Average & roof & beam & frame & bicycle & exer. & microwave & printer & oven & mailbox & washbasin & partition & piano & countertop & drawer & carpet \\
\midrule
Baseline (rep.) & 7.12 & 0.00 & 0.00 & 0.00 & 20.98 & 0.65 & 16.94 & 40.00 & 13.08 & 0.16 & 0.62 & 2.54 & 5.23 & 0.33 & 1.25 & 4.95 \\
DROcc & 9.66 & 0.00 & 0.00 & 0.00 & 23.58 & 2.72 & 23.02 & 32.70 & 8.43 & \textbf{12.41} & 0.70 & 13.00 & 11.38 & 8.12 & 1.53 & 7.30 \\
b.CLIP-ResNet & 3.44 & 0.00 & 0.00 & 0.00 & 0.00 & 0.00 & 13.98 & 14.72 & 7.21 & 0.00 & 1.09 & 0.00 & 4.55 & 1.19 & 0.75 & 8.04 \\
b.CLIP-ViT & 9.86 & 0.00 & 0.00 & 0.00 & 28.30 & 15.41 & 22.19 & 32.05 & 8.90 & 0.00 & 1.35 & 0.27 & 28.03 & \textbf{9.20} & \textbf{2.15} & 0.00 \\
b.BLIP2 & 15.82 & 0.00 & 0.00 & \textbf{0.74} & \textbf{52.16} & \textbf{32.67} & 30.06 & \textbf{43.14} & 12.45 & 1.83 & \textbf{4.55} & 3.18 & \textbf{50.39} & 4.72 & 1.29 & 0.13 \\
b.DINOv2 & \textbf{17.28} & 0.00 & 0.00 & 0.00 & 49.00 & 19.03 & \textbf{39.46} & 41.73 & \textbf{20.35} & 5.82 & 2.25 & \textbf{22.96} & 42.76 & 7.91 & 2.08 & 5.91 \\
\bottomrule
\end{tabular}
}
\end{table*}

% \begin{table*}[t]
% \centering
% \caption{IoU on the 15 least frequent classes used for long-tail analysis. Values are percentages. ``exer.'' denotes exercise equipment. These categories follow the rare-class grouping used in DROcc~\cite{fang2026drocc}.}
% \label{tab:longtail}
% \resizebox{\linewidth}{!}{
% \begin{tabular}{l|ccccccccccccccc}
% \toprule
% Method & roof & beam & frame & bicycle & exer. & microwave & printer & oven & mailbox & washbasin & partition & piano & countertop & drawer & carpet \\
% \midrule
% Baseline (rep.) & 0.00 & 0.00 & 0.00 & 20.98 & 0.65 & 16.94 & 40.00 & 13.08 & 0.16 & 0.62 & 2.54 & 5.23 & 0.33 & 1.25 & 4.95 \\
% DROcc & 0.00 & 0.00 & 0.00 & 23.58 & 2.72 & 23.02 & 32.70 & 8.43 & \textbf{12.41} & 0.70 & 13.00 & 11.38 & 8.12 & 1.53 & 7.30 \\
% b.CLIP-ResNet & 0.00 & 0.00 & 0.00 & 0.00 & 0.00 & 13.98 & 14.72 & 7.21 & 0.00 & 1.09 & 0.00 & 4.55 & 1.19 & 0.75 & 8.04 \\
% b.CLIP-ViT & 0.00 & 0.00 & 0.00 & 28.30 & 15.41 & 22.19 & 32.05 & 8.90 & 0.00 & 1.35 & 0.27 & 28.03 & \textbf{9.20} & \textbf{2.15} & 0.00 \\
% b.BLIP2 & 0.00 & 0.00 & \textbf{0.74} & \textbf{52.16} & \textbf{32.67} & 30.06 & \textbf{43.14} & 12.45 & 1.83 & \textbf{4.55} & 3.18 & \textbf{50.39} & 4.72 & 1.29 & 0.13 \\
% b.DINOv2 & 0.00 & 0.00 & 0.00 & 49.00 & 19.03 & \textbf{39.46} & 41.73 & \textbf{20.35} & 5.82 & 2.25 & \textbf{22.96} & 42.76 & 7.91 & 2.08 & 5.91 \\
% \bottomrule
% \end{tabular}
% }
% \end{table*}

\subsection{Long-Tail Category Discussion}
\label{sec:longtail_discussion}

% Table~\ref{tab:longtail} shifts the analysis from common categories to the 15 least frequent classes. The first observation is that some categories remain nearly unsolved: roof and beam are zero for every method, and frame is almost always zero. These failures are not caused by a single backbone; they reflect the limited supervision and sparse visibility of rare indoor labels.
Table~\ref{tab:longtail} shifts the analysis from common categories to the 15 least frequent classes. The first observation is that some categories remain nearly unsolved: roof and beam are zero for every method, and frame is almost always zero. These failures are not caused by a single backbone. They combine three factors: severe data imbalance, weak geometric observability from ego-centric RGB-D views, and limited feature resolution after projection. Roof and beam are structural labels that are often only partially visible or outside the dominant camera frustum, so stronger 2D descriptors cannot reliably supply evidence for them. Frame is usually thin and boundary-like, making it sensitive to voxel resolution and bilinear sampling. This suggests a category-specific transition point: for these labels, the 2D encoder is no longer the only bottleneck, and further gains likely require visibility-aware lifting, rare-class supervision, or higher-resolution geometric modeling.
The second observation is that stronger backbones still help many rare objects. BLIP2 is strongest on bicycle, exercise equipment, printer, washbasin, and piano. DINOv2 is strongest on microwave, oven, and partition, and is close to BLIP2 on bicycle and printer. CLIP-ViT gives large gains over CLIP-ResNet on bicycle, exercise equipment, piano, countertop, and drawer. However, the table also shows that rare-class improvement is uneven: a method can be strong on object-like rare classes while remaining weak on structural rare labels. Backbone replacement therefore improves long-tail representation, but additional sampling, loss reweighting, or rare-class modeling is still needed.

\begin{table}[t]
\centering
\caption{Resource comparison. Training time is reported in hours, memory in GiB, inference time in seconds per instance, and parameters in millions.}
\label{tab:resources}
% \resizebox{\linewidth}{!}{
\begin{tabular}{l|ccccc}
\toprule
Method & Train time & Train mem. & Test time & Test mem. & Params \\
\midrule
Baseline & 38.03 & 19.86 & 2.65 & 7.15 & 751.28 \\
DROcc & 33.83 & 6.74 & 2.59 & 2.73 & 114.38 \\
b.CLIP-ResNet & 40.06 & 19.18 & 2.82 & 7.39 & 751.28 \\
b.CLIP-ViT & 35.98 & 19.73 & 2.70 & 7.83 & 787.98 \\
b.BLIP2 & 49.90 & 21.71 & 3.75 & 13.52 & 1718.77 \\
b.DINOv2 & 39.02 & 18.84 & 3.11 & 10.63 & 1034.62 \\
\bottomrule
\end{tabular}
% }
\end{table}

\subsection{Efficiency and Model Cost}
\label{sec:efficiency}

Table~\ref{tab:resources} gives the resource side of the comparison. BLIP2 has the largest cost, with 1718.77M parameters, 49.90 training hours, and 13.52 GiB test memory. DINOv2 is lighter than BLIP2 but still substantially larger than the baseline in parameter count. CLIP-ViT has a modest parameter increase over the baseline and nearly the same runtime profile. DROcc is much smaller and cheaper than the baseline because it redesigns the fusion architecture, while our backbone variants keep the baseline 3D pipeline and only alter the image encoder. These numbers clarify the trade-off: DINOv2 and BLIP2 bring the strongest accuracy, but they also increase model size and, for BLIP2 especially, inference memory.

\begin{figure*}[ht]
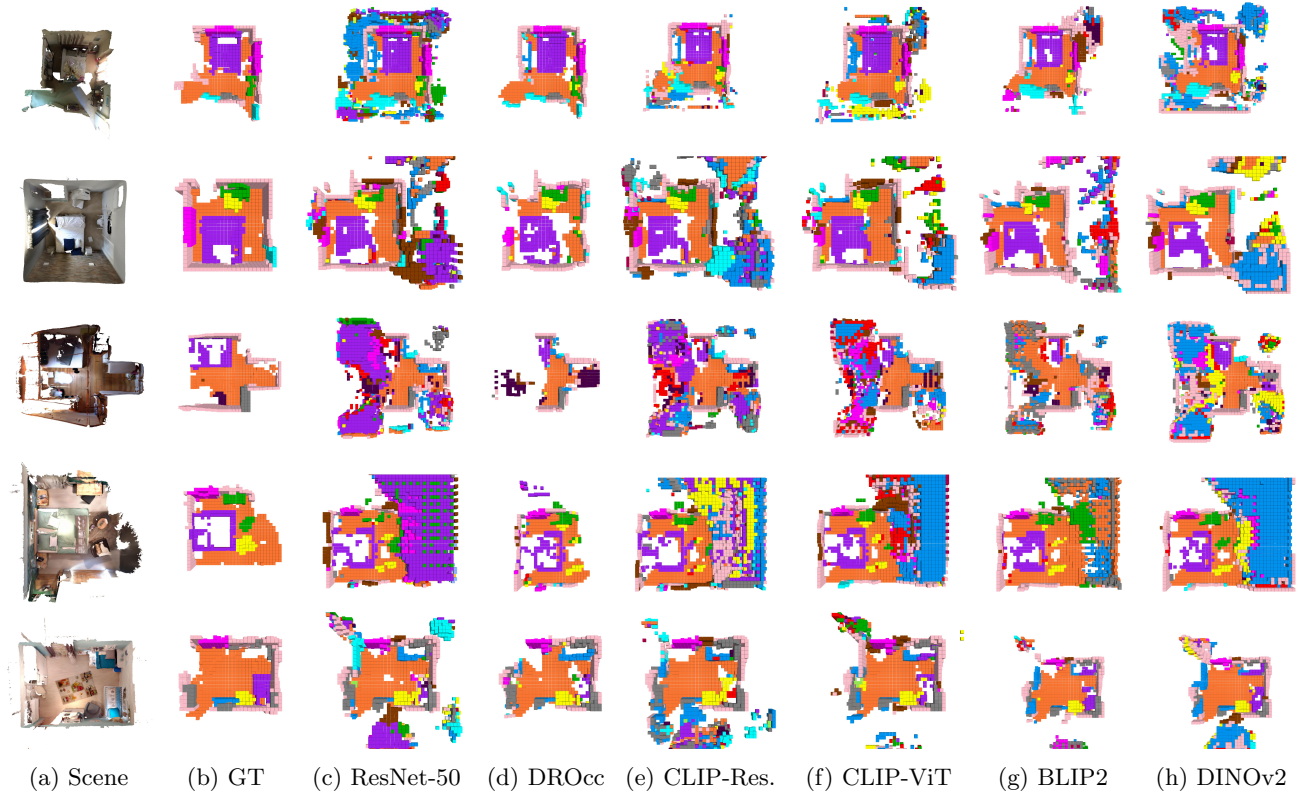
  % htbp：浮动位置参数（here, top, bottom, page）
    \centering       % 整体居中
        % 子图1
    \begin{subfigure}[b]{0.1150\textwidth}  % [b]底部对齐，宽度占文本宽度的30%
        \centering
        \includegraphics[width=\textwidth]{./vis/1.pdf}  % 图片宽度占子图宽度
        \caption{Scene}  % 子图标题
        \label{fig:sub1}    % 子图标签（引用用）
    \end{subfigure}
    \hfill  % 子图间填充空白（平均分布）
    % 子图2
    \begin{subfigure}[b]{0.0978\textwidth}
        \centering
        \includegraphics[width=\textwidth]{./vis/2.pdf}
        \caption{GT}
        \label{fig:sub2}
    \end{subfigure}
    \hfill
   % 子图3
    \begin{subfigure}[b]{0.1361\textwidth}
        \centering
        \includegraphics[width=\textwidth]{./vis/3.pdf}
        \caption{ResNet-50}
        \label{fig:sub3}
    \end{subfigure}
    \hfill
    % 子图4
    \begin{subfigure}[b]{0.0952\textwidth}
        \centering
        \includegraphics[width=\textwidth]{./vis/4.pdf}
        \caption{DROcc}
        \label{fig:sub4}
    \end{subfigure}
    \hfill
    % 子图5
    \begin{subfigure}[b]{0.1291\textwidth}
        \centering
        \includegraphics[width=\textwidth]{./vis/5.pdf}
        \caption{CLIP-Res.}
        \label{fig:sub5}
    \end{subfigure}
    \hfill
    % 子图6
    \begin{subfigure}[b]{0.1315\textwidth}  % [b]底部对齐，宽度占文本宽度的30%
        \centering
        \includegraphics[width=\textwidth]{./vis/6.pdf}  % 图片宽度占子图宽度
        \caption{CLIP-ViT}  % 子图标题
        \label{fig:sub6}    % 子图标签（引用用）
    \end{subfigure}
    \hfill  % 子图间填充空白（平均分布）
    % 子图7
    \begin{subfigure}[b]{0.1231\textwidth}
        \centering
        \includegraphics[width=\textwidth]{./vis/7.pdf}
        \caption{BLIP2}
        \label{fig:sub7}
    \end{subfigure}
       \hfill  % 子图间填充空白（平均分布）
    % 子图8
    \begin{subfigure}[b]{0.1223\textwidth}
        \centering
        \includegraphics[width=\textwidth]{./vis/8.pdf}
        \caption{DINOv2}
        \label{fig:sub8}
    \end{subfigure}

    \vspace{3pt}
   
    \caption{Qualitative comparison of semantic occupancy predictions. Columns show indoor scene input, ground truth, EmbodiedScan baseline, DROcc, CLIP-ResNet, CLIP-ViT, BLIP2, and DINOv2. CLIP-ResNet produces noisier and less complete object layouts, CLIP-ViT recovers more coherent foreground regions, and BLIP2 and DINOv2 recover richer semantic occupancy structures. DINOv2 is generally closer to the ground-truth room structure, whereas BLIP2 often preserves strong object-level responses.}  % 总标题
    \label{fig:occ_vis}  % 总图标签
\end{figure*}

\subsection{Qualitative Results}
\label{sec:qualitative}

Figure~\ref{fig:occ_vis} shows the same trend qualitatively. CLIP-ResNet produces fragmented regions and unstable foreground objects, matching its low mIoU. CLIP-ViT fills in more of the foreground and is less noisy, although it still misses some category distinctions. BLIP2 often recovers more object responses than the CLIP variants, but it can also leave scattered predictions near scene boundaries. DINOv2 gives cleaner room structure in several examples, especially on large surfaces and frequently observed objects.

These examples also explain why image-level semantic strength is not the whole story. After camera projection and view averaging, a useful feature must remain locally aligned and semantically separable. DINOv2 and BLIP2 meet this requirement better than the two CLIP variants, but not in the same way: DINOv2 looks more stable for layout and dense surfaces, whereas BLIP2 is still competitive for object-centered responses. CLIP-ViT confirms that dense token features help CLIP transfer, while its remaining gap shows that architecture and pretraining objective have to be considered together.

\subsection{Discussion}
\label{sec:discussion}

The experiments lead to three conclusions. First, the image backbone should be treated as a first-order experimental variable in RGB-D occupancy prediction. A change in the 2D encoder alone can produce more than 13 mIoU points of difference under the same 3D pipeline. Second, DINOv2 is the strongest backbone in our controlled study, suggesting that self-supervised dense visual representations are well aligned with projection-based voxel prediction. Third, vision-language representations remain useful but require careful interpretation: BLIP2 is close to DINOv2 overall and wins several object-centric categories, CLIP-ViT substantially improves over CLIP-ResNet, and the tested CLIP-ResNet variant is much weaker for dense semantic occupancy.

%% file: section/conclusion.tex
% !TEX root = ./main.tex

\section{Conclusion}
\label{sec:conclusion}

This paper studies the influence of image backbones on EmbodiedScan semantic occupancy prediction. Under a controlled RGB-D pipeline, we replace only the 2D image encoder and keep the depth branch, voxel projection, fusion neck, occupancy head, losses, and training protocol fixed. The results show that backbone choice has a large impact: DINOv2 achieves the best mIoU of 30.55\%, BLIP2 reaches 29.49\%, CLIP-ViT reaches 24.33\%, and CLIP-ResNet reaches 17.41\%. Compared with the original EmbodiedScan baseline, our reproduced baseline, DROcc(SwinU), and DROcc, the stronger backbone variants show that a front-end representation change can exceed several task-specific architectural improvements. The weaker CLIP-ResNet result also shows that pretraining alone does not guarantee better dense occupancy features.

The category-level analysis gives a more detailed conclusion. DINOv2 is strong on many layout and structural classes, while BLIP2 remains competitive on several object-centered classes. CLIP-ViT improves over CLIP-ResNet throughout the common-category table, demonstrating that dense token features matter within CLIP-style pretraining. 
For the 15 least frequent categories, BLIP2 and DINOv2 recover several rare objects more effectively, but roof, beam, and frame remain nearly unsolved. This does not weaken the backbone finding; rather, it marks where backbone improvement has category-specific diminishing returns. When labels are extremely rare, weakly observed, or thinner than the effective voxel/image sampling resolution, the limiting factor shifts from visual representation quality to visibility modeling, supervision balance, and geometric resolution. These findings indicate that image backbone selection should be explicitly reported in embodied occupancy studies. Future work should combine stronger visual encoders with visibility-aware lifting, long-tail learning, improved RGB-D fusion, and decoders designed for sparse indoor objects.
% For the 15 least frequent categories, BLIP2 and DINOv2 recover several rare objects more effectively, but roof, beam, and frame remain nearly unsolved. These findings indicate that image backbone selection should be explicitly reported in embodied occupancy studies. Future work should combine stronger visual encoders with long-tail learning, improved RGB-D fusion, and decoders designed for sparse indoor objects.